\documentclass{article}

\usepackage{arxiv}
\usepackage{subfigure}
\usepackage{graphicx}
\usepackage[utf8]{inputenc} % allow utf-8 input
\usepackage[T1]{fontenc}    % use 8-bit T1 fonts
\usepackage{hyperref}       % hyperlinks
\usepackage{url}            % simple URL typesetting
\usepackage{booktabs}       % professional-quality tables
\usepackage{amsfonts}       % blackboard math symbols
\usepackage{nicefrac}       % compact symbols for 1/2, etc.
\usepackage{microtype}      % microtypography
\usepackage{lipsum}

\usepackage{amsmath}
\usepackage{float}
\usepackage{multirow}
\usepackage{gensymb}

\title{Mapping Road Safety Features from Streetview Imagery: A Deep Learning Approach}

\author{
  Arpan M.~Sainju\thanks{Use footnote for providing further
    information about author (webpage, alternative
    address)---\emph{not} for acknowledging funding agencies.} \\
  Department of Computer Science\\
  University of Alabama\\
  Tuscaloosa, AL 35487 \\
  \texttt{asainju@crimson.ua.edu} \\
   \And
 Zhe Jiang \\
  Department of Computer Science\\
  University of Alabama\\
  Tuscaloosa, AL 35487 \\
  \texttt{zjiang@cs.ua.edu} \\
}

\begin{document}
\maketitle

\begin{abstract}
Each year, around 6 million car accidents occur in the U.S. on average. Road safety features (e.g., concrete barriers, metal crash barriers, rumble strips) play an important role in preventing or mitigating vehicle crashes. Accurate maps of road safety features is an important component of safety management systems for federal or state transportation agencies, helping traffic engineers identify locations to invest on safety infrastructure. In current practice, mapping road safety features is largely done manually (e.g., observations on the road or visual interpretation of streetview imagery), which is both expensive and time consuming. In this paper, we propose a deep learning approach to automatically map road safety features from  streetview imagery. Unlike existing Convolutional Neural Networks (CNNs) that classify each image individually, we propose to further add Recurrent Neural Network (Long Short Term Memory) to capture  geographic context of images (spatial autocorrelation effect along linear road network paths). Evaluations on real world streetview imagery show that our proposed model outperforms several baseline methods. 
\end{abstract}

% keywords can be removed

\section{Introduction}
\label{sect:intro}
\begin{figure*}
	\centering
	\subfigure{\includegraphics[width=1.8in]{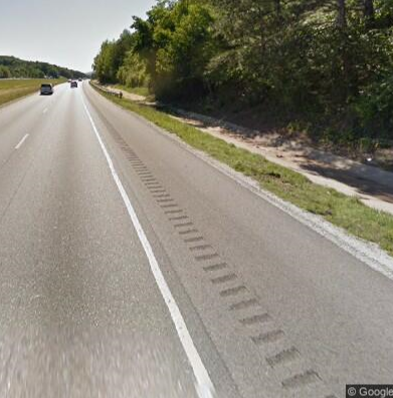}}
	\subfigure{\includegraphics[width=1.8in]{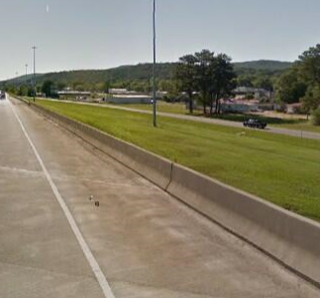}}
	\subfigure{\includegraphics[width=1.8in]{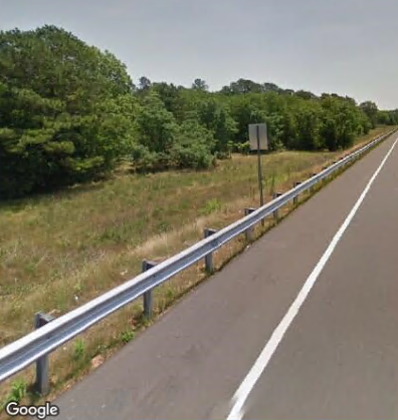}}

	\caption{Three common classes of road safety features on Google Streetview Imagery. (a) Rumble strips (b) Concrete barrier (c) Metal crash barrier}
	\label{fig:barriers}
\end{figure*}

Every year, around 6 million car accidents occur in the U.S. on average~\cite{accident}. Traffic safety has long been an important societal issue. In order to avoid or mitigate vehicle crashes, traffic engineers place roadside barriers to prevent out of control vehicles from diverting off the roads and hitting the roadside hazards. Such road safety features can also prevent vehicles from crossing into the path of other vehicles. During winter season, vehicles can become more difficult to control on icy and slippery road surface, particularly when the vehicle speed is high. Barriers on the roadside can act as a safety precaution in such cases. Other safety features such as rumble strips help alert inattentive drivers who are deviating from their lanes.  Figure~\ref{fig:barriers} shows three different common type of road safety features, rumble strip, concrete barrier, and metal crash barrier. Federal, state and local governments spent several hundred billion dollars each year on transportation infrastructure development and maintenance~\cite{budget}. Mapping safety features along road networks can play a crucial role in managing and maintaining road safety infrastructures. Traffic engineers can use the detailed safety feature map to identify locations where new safety infrastructure should be invested.

% Rumble Strip : https://www.ayresassociates.com/rumble-strips-help-keep-drivers-safe/
% Concrete: https://48barriers.com/interesting-facts-concrete-jersey-barriers/
% Metal Crash : https://www.alibaba.com/product-detail/Metal-Beam-Crash-Barrier_50028860028.html

In current practice, mapping road safety features are mostly done manually by well-trained traffic engineers driving through road networks or visually interpreting streetview images. A {\bf streetview image} is a geo-referenced image taken at a specific location on the ground. One common example is Google Streetview Imagery collected by vehicles equipped with GPS and cameras driving along streets on road networks. However, such a manual process is both expensive and time-consuming. Given the large amount of information to collect, the cost of these approaches quickly become prohibitive. 

The focus of this paper is to develop a deep learning algorithm that can automatically map road safety features from streetview imagery. The results can be used by the transportation agencies in management and maintenance of road safety infrastructures, as well as planning the investment on new infrastructures. Specifically, we can utilize a small set of manually labeled imagery (whose road safety features are visually inspected) to learn a classification model. Then the model can be used to classify safety feature types on a large number of unlabeled imagery along the road network.  

However, mapping the road safety features based on streetview imagery poses several unique challenges. First, streetview images are not independent and identically distributed along a road network. In contrast, the safety feature types of consecutive images along a same road network path often resemble each other (also called the spatial autocorrelation effect). Second, the spatial scales of road safety features may vary across different class categories. For example, concrete barrier is often very long (e.g., miles). In contrast, metal crash barriers are much shorter (e.g., hundred meters). Third, individual images may be imperfect due to some noise or obstacles.  For example, a safety feature can be blocked by a large vehicle and thus become invisible in an image. 

To address these challenges, we propose a deep learning model based on both convolutional and recurrent units. We use covolutional neural network (CNN) model to extract semantic features from individual images. We also use a recurrent neural network, Long Short-Term Memory (LSTM), to model spatial sequential structure on extracted features from consecutive images along a road network path (the spatial autocorrelation effect). The integration of CNN and LSTM enables our deep learning model to utilize not only the content of individual images but also the geographic context between images.
Evaluations on real world streetview images collected from highways in Alabama show that our approaches outperform several baseline methods in classification performance.

In summary, the contributions of this paper are listed below: 
\begin{itemize}
    \item To the best of our knowledge, we are the first to explore a deep learning approach on Google Streetview imagery for road safety feature mapping. 
    \item We propose to use integrated deep learning models that combine CNN and LSTM. The integrated models can utilize not only the content of individual images but also the spatial sequential structure between images. 
    \item We compare our approaches with several baseline methods on two real world streetview imagery datasets collected in Alabama. We achieve 3 and 5 percent improvement in F-score over the best baseline method on two different test datasets.
    \item We perform a case study of mapping road safety features with 69,500 streetview imageries over all the major highways in the entire state of Alabama. 
\end{itemize}

The outline of the paper is as follows. Section ~\ref{sec:relatedworks} discusses some of the related works. Section~\ref{sec:defination} formally defines the problem. Section~\ref{sec:approach} introduces the approaches. Section~\ref{sec:experiment} summarizes the results of our experimental evaluation on two real world datasets as well as discusses the case study of mapping road safety features over all the major highways in Alabama. Section~\ref{sec:conclusion} concludes the paper with discussions on future works.

\section{Related Works}\label{sec:relatedworks}
In this section, we briefly review the relevant research on transportation safety and deep learning techniques for spatial and spatiotemporal data.

\subsection{Transportation Safety}
Related work in transportation safety often focuses on analyzing the protective effect of different road safety features (e.g., roadside barriers)~\cite{karim2012assessment,roque2013observations,zou2014effectiveness}. For example, studies in~\cite{bambach2013protective,jama2011characteristics,grzebieta2013motorcyclist} quantify the protective effect of barriers with regards to motorcyclist injury.  Work in~\cite{bryden1986traffic} analyzes the performance of roadside barriers related to vehicle size and type. \cite{vieira2008development} studies how to increase the effectiveness of the roadside barriers in safety protection. For example, studies found that concrete barriers can hold high-energy truck crash, but can also cause more fatalities. Some recent work focuses on developing energy absorbing barrier~\cite{schmidt2013development}. Beside the protective effect, other studies on roadside barriers focus on the impact on mitigating near-road air pollution~\cite{tong2016roadside}. The study on effect of solid barriers on dispersion of roadway emissions in~\cite{schulte2014effects,hagler2011model} shows that roadside barriers is one of the most practical mitigation methods. There are also works that analyze spatial patterns from traffic accident event locations such as network hotspots and colocation patterns~\cite{romano2017visualizing,sainju2017grid,sainju2018parallel}.  \cite{jiang2016identifying} proposes efficient algorithms to identify primary corridors from cyclists' GPS trajectories on urban road networks to study riding behaviors for safety issues. \cite{musaev2018detection} shows techniques to detect coarse scale hotspots of road failure events through geo-tagged tweets from social media.

%Google street view images 
Recently, researchers have used Google Streetview imagery along the road network for traffic sign detection for roadway inventory management~\cite{balali2015multi,balali2015detection,tsai2016traffic}. Other works use streetview imagery to estimate the demographic makeup of neighborhoods~\cite{gebru2017using}, to assess street-level greenness in an urban area~\cite{li2015assessing}. To the best of our knowledge, there is little research on utilizing streetview imagery to automatically map road safety features. 

\subsection{Deep Learning for Spatio-Temporal Data}
In recent years, deep learning techniques have shown great growth in the field of spatiotemporal data mining~\cite{shekhar2015spatiotemporal,jiang2018survey}. One common approach is to integrate deep convolutional neural networks (CNN) with recurrent neural networks such as Long Short-Term Memory (LSTM). The CNN component can be used to model spatial dependency structure in one temporal snapshot, while the LSTM component can be used to model temporal dynamics between different snapshots. For example, \cite{Zhang_2017_ICCV} uses fully convolutional networks with LSTM to estimate vehicle count maps based on city cameras.
\cite{yao2018deep} uses CNN-LSTM model together with multi-view learning to predict taxi demand. 
 \cite{wu2016short} uses a one-dimensional CNN to capture spatial features of traffic flow and two LSTM models to capture the short-term variability and periodicities of traffic flow. 
\cite{yao2018modeling} addresses the traffic prediction problem with a new spatiotemporal model. It uses a flow gating mechanism to learn the dynamic similarity between locations, and uses a periodically shifted attention mechanism to handle long-term periodic temporal shifting. \cite{yuan2018hetero,yuan2017predicting} researches on better traffic accident prediction to improve transportation and public safety. In these existing works, LSTM model is often used to model temporal dynamics between multiple spatial snapshots. The difference from our work in this paper is that we use LSTM to capture linear spatial sequential structure between consecutive images along a road network path.

\section{Problem Description}\label{sec:defination}

In this section, we discuss some basic concepts and describe our problem. 

%here, i like the math symbols to show formal, but since these symbols are not used later, I just removed them for simplicity
\emph{Road network:} A road network is a network whose nodes are road intersections, and whose edges are road segments. At the same time, a road network is also a spatial network whose nodes are spatial points and whose edges are spatial line strings. In other words, a road network has both graph properties and geometric properties. 

\emph{Streetview imagery:} Streetview imagery is a sequence of geo-referenced images whose locations are embedded on road network edges (in the form of line strings). The imagery is collected through driving a vehicle equipped with GPS and camera, so that each image can be geo-referenced based on the GPS time stamp. In this paper, we used Google Streetview API to select imagery at a regular spatial interval of 20 meters.

\emph{Road safety feature:} A road safety feature is defined as the measure or infrastructure placed on a road to improve safety. We consider three most common safety features: rumble strips, concrete barrier and metal crash barrier. Figure ~\ref{fig:barriers} shows examples of the three safety features from Google Streetview imagery.
\begin{itemize}
    \item \emph{Rumble Strips:}
Rumble strips (Figure~\ref{fig:barriers}(a)) are milled grooves or rows of raised pavement markers placed perpendicular to the direction of travel, or a continuous sinusoidal pattern milled longitudinal to the direction of travel. It creates a vibration and rumbling sound transmitted through the wheels into the vehicle interior which can alert the drivers who have drifted from their lanes. 
    \item \emph{Concrete barrier:}
Concrete barrier (Figure~\ref{fig:barriers}(b)) is a rigid barrier. It is easy to maintain. This type of barrier is often used on roads where traffic in opposing direction is flowing in close proximity due to lack of space. 
    \item \emph{Metal crash barrier:}
Metal crash barrier (Figure~\ref{fig:barriers}(c)), also known as guardrails, is usually made from steel beams or rails. It ensures minimum damage to the vehicle and its occupants by absorbing the impact energy of the colliding vehicle.  It can also act as a good visual guide during night time for the driver to maintain their lane position.
\end{itemize}

\emph{Problem Definition:}
Given a road network with geo-referenced streetview imagery sampled at an equal distance interval, as well as a small collection of labeled imagery sequences (each image has three binary class labels corresponding to the existence of rumble strips, concrete barrier, and metal crash barrier respectively), the road safety feature mapping problem aims to learn a classification model that can predict the labels for all unlabeled images on the road network.
Since each image may contain multiple types of road safety features at the same time, our problem is a multi-label classification problem. 
% Suppose we have a set of continuous sequences of street view images along the road segment, $\mathbf{S}=\{\mathbf{s}_1, \mathbf{s}_2,..,\mathbf{s}_k\}$ along with the  ground truth label for each images in the sequence $G$ and target set of continuous sequence $S_{target}$ of street view images along different road segment without the ground truth label. Our goal is to predict the roadside barrier along the target road segment $S_{target}$.

% {\bf Input:}
% \begin{itemize}
%     \item set of sequence of continuous street view images, $S$
%     \item Ground truth label for each image in the sequences, $G$
%     \item set of sequence of continuous street view images without ground truth, $S_{target}$
% \end{itemize}

% {\bf Output:}
% \begin{itemize}
%     \item Predicted roadside barrier labels for street view images in target road segment $S_{target}$.
% \end{itemize}

% {\bf Objective:}
% \begin{itemize}
%     \item Train the model using the input sequence $S$ and its ground truth label $G$
% \end{itemize}

% {\bf Constraints:}
% \begin{itemize}
%     \item Minimize the training loss
% \end{itemize}

 \begin{figure*}[h]
	\includegraphics[width=6.5in]{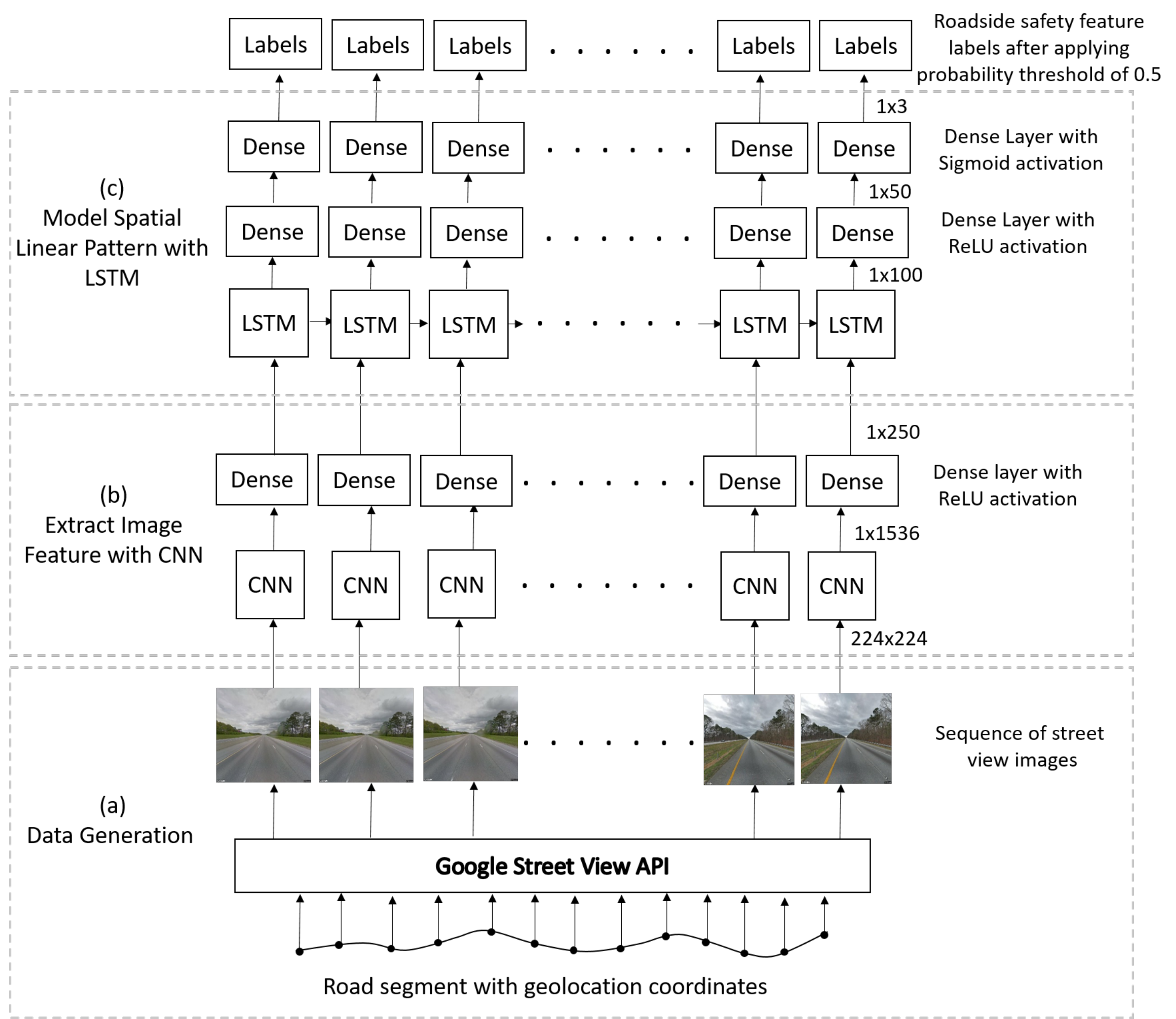}
	\caption{Overall framework of our deep learning approach}
	\label{fig:architecture}
\end{figure*}

\section{Proposed Approach}\label{sec:approach}
In this section, we introduce our proposed deep learning approaches.  Figure~\ref{fig:architecture} illustrates the overall framework of our proposed models. The bottom component shows the data collection process. We sample a number of spatial points along road network edges at an equal distance interval (e.g., 20 meters), and then use Google Streetview API to download geo-referenced imagery at those point locations. We fixed the distance interval of 20 meters because the average length of some road safety features such as metal crash barrier is only a few hundred meters. If we select a higher distance interval, there may not exist enough images for short extent barrier such as metal crash barrier. Although lower distance interval can provide  fine-grained dataset, it increases the number of streetview images to be downloaded which incurs extra cost. The middle component of our proposed models is based on retrained CNN model to extract low dimensional features from individual images. The last component is  LSTM layer. In contrast to existing works, our LSTM does not capture temporal dynamics between different spatial snapshots, but represents spatial sequential pattern between consecutive imagery along road network edges. 

\subsection{Extract Image Feature with CNN}
Convolutional Neural Network (CNN) was developed mainly for image classification. CNN introduces the concept of parameter sharing which allows the model to learn less number of parameters in comparison to regular neural network. Similar to regular neural networks, CNN also consists of a sequence of layers. We briefly describe each layer in CNN below. 

\emph{Input Layer:} Input Layer holds the raw pixel color values (RGB) of the images. Usually, the pixel values are normalized to stabilize the learning process and dramatically reduce the number of training epochs required to train deep learning models.  

\emph{Convolution Layer:} Convolutional layer transforms the input using convolution operation. A convolution operation is element-wise multiplication of a pixel and its neighborhood pixels color value (RGB) by a matrix. It is also known as convolution filter. Different filters are used to convolve around all the pixels in an image. Filters like horizontal and vertical edge detecting filter can  extract the linear feature from the image. Other complicated filters such as sobel filters can extract non-linear edges. In CNNs, filters are not defined, they are learned during the training process. By stacking layers of convolutions on top of each other, we can get more abstract and in-depth information from a CNN.

\emph{ReLU Layer:} ReLU stands for Rectified Linear Unit, which is a type of activation function commonly used in neural networks. Activation functions are applied to introduce non-linear properties to the network. The function returns 0 if it receives any negative input. However, for any positive value $x$, the function returns the same value back. So, it can be written as $f(x)=max(0,x)$. ReLU activation function is computationally less expensive as there is no complicated math, which can reduce the model training time.

\emph{Pooling layer:} The function of pooling layer is to reduce the spatial size of the input. It is also known as downsampling layer. Pooling layer can reduce the number of parameters and computation in the network.  It applies a filter (usually of size 2x2) to the input volume. Pooling filters can be based on different operations such as max, min or average. The most common one is max filter which extracts the max value from the filter region. 
%image for pooling?

\emph{Fully Connected Layer:}
Fully connected (Dense) layer takes an input volume (output of activation function) and outputs a N-dimensional vector. Similar to regular neural networks, neurons in a fully connected layer have full connections to all activations in the previous layer. 

For our proposed models, we use the current state-of-art Inception-ResnetV2~\cite{szegedy2017inception} model to extract features from the streetview images. We use the keras implementation of Incpetion-ResnetV2 pre-trained on ImageNet \cite{deng2009imagenet} dataset with 1000 classes. Incpetion-ResnetV2 combines the idea of residual connections to inception architecture. In residual connection, each layer feeds into the next layer and directly into the layers about few hops away. Residual connections are important for very deep architecture. When deeper networks starts converging, the accuracy can saturate at a point and eventually degrade. Residual connections are designed to overcome this degrading problem. As the Inception-v4 network is very deep with around 200 layers, combining Inception architecture with residual connections can be beneficial.  

We remove the final dense layer with softmax activation function  because the network has been pretrained to classify 1000 classes. In our work, we only have 3 classes: rumble strips, concrete barrier, and metal crash barrier. Next, we add a dense layer with 250 nodes after the last average pooling layer (with 1,536 nodes). We reduce the feature dimension because we are classifying our dataset into a lower number of classes than the pretrained model. Finally, we add a dense layer with 3 nodes with a sigmoid activation function so that each node provides a probability value for one class label. 

As shown in Figure~\ref{fig:architecture}, we retrain the CNN model using our streetview dataset. The input to the CNN model is a set of  224x224 streetview images. After retraining, we extract the output of dense layer with 250 nodes to get a feature vector of 250 dimensions for each image in the sequence.  We then create a set of feature sequences to feed into the LSTM model. 

\subsection{Model Spatial Linear Pattern with LSTM}
In order to model spatial linear (sequential) structure along a road network path, we used the LSTM model on a sequence of image features extracted by the CNN model. LSTM is a type of recurrent neural networks that uses gating functions to avoid the exploding and vanishing gradient issues. The gate function can help a model to memorize the state of previous units in a sequence.  Such recurrent structure is well-suitable to capture the spatial autocorrelation effect across consecutive images. According to the first law of geography: "everything is related to everything else, but near things are more related than distant things." For example, concrete barriers are often very long spanning over several miles. Metal crash barriers, in contrast, have a shorter spatial scale within a few hundred meters. 
 \begin{figure}[h]
 \centering
	\includegraphics[width=2.8in]{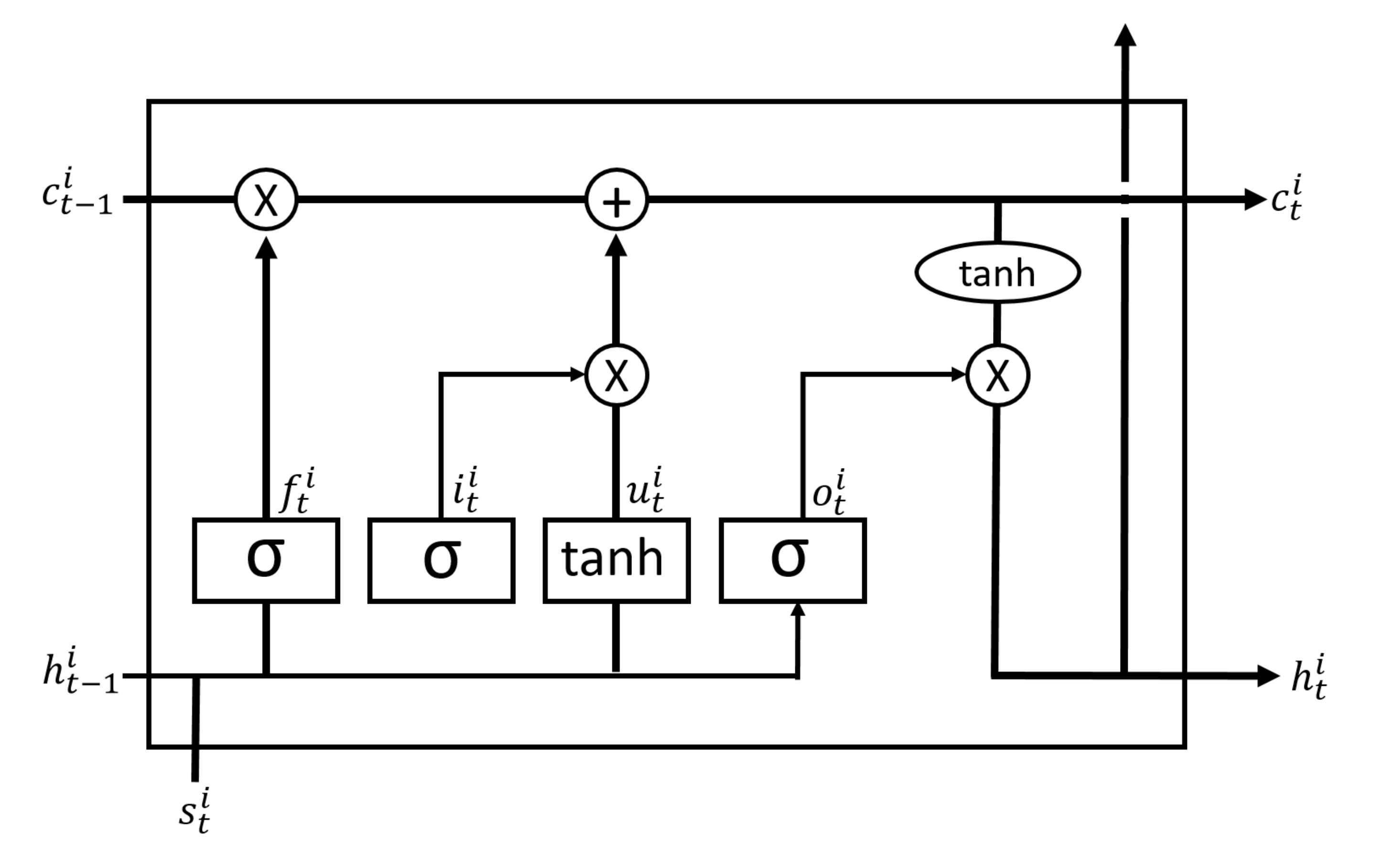}
	\caption{LSTM Unit}
	\label{fig:lstm}
\end{figure}

LSTM models a sequential structure by maintaining a sequence of memory cells ($c_{t}$ with $t$ as the spatial location index). In each spatial location $t$, LSTM takes an input feature $s_{t}^{i}$, hidden state $h_{t-1}$ and cell state $c_{t-1}$. Figure~\ref{fig:lstm} shows a LSTM unit with a cell state($c_{t}$) and three different gates: input gate, output gate and forget gate. The forget gate ($f_{t}$) decides how much information from a previous cell unit is ignored before coming to the next cell. The input gate ($i_{t}$) decides how much contribution an input feature vector makes to the current cell state. Finally, the output gate ($o_{t}$) decides what the current LSTM unit is going to output (current cell state $c_{t}$ and current hidden state $h_{t}$) based on the cell state. The LSTM transaction equations are as follows,
\begin{equation} \label{finalPrediction}
\begin{gathered}
f_{t}^{i} = \sigma (W_{f}h^{i}_{t-1}+ U_{f}s_{t}^{i}+ b_{f})\\
i_{t}^{i} = \sigma (W_{i}h^{i}_{t-1}+ U_{i}s_{t}^{i}+ b_{i})\\
o_{t}^{i} = \sigma (W_{o}h^{i}_{t-1}+ U_{o}s_{t}^{i}+ b_{o}) \\
u_{t}^{i} = tanh W_{u}h^{i}_{t-1}+ U_{u}x_{t}^{i}+ b_{u}) \\
c_{t}^{i} = f_{t} * c^{i}_{t-1} + i_{t}^{i} * u_{t} \\
h_{t}^{i} = o_{t} * tanh (c_{t}^{i}) 
\end{gathered}
\end{equation}
where $\sigma$ denotes the sigmoid activation function, $tanh$ is hyperbolic tangent function and $*$ denotes element-wise product. $W$ and  $U$ denote model parameters. As Figure~\ref{fig:architecture}(c) shows, our LSTM model consists of 4 hidden layers. The first layer is an LSTM layer with an output dimension of 100 units. The second layer is a 20\% dropout layer. The third layer is a dense layer with 50 nodes. 
Our problem is a multi-label classification problem because each image may contain multiple types of road safety features at the same time.
 So, we implemented two design decision for the last layer to handle multi-label classification issue which will be discussed in detail in  subsection 4.3. 

\subsection{Multi-label Classification}
To model multi-label classification, we propose two approaches. First approach involves training a shared CNN-LSTM model for all three class label together. Second approach involves training separate CNN-LSTM models for each class label independently.

In first approach, the last layer in the model is a sigmoid transformation layer with 3 nodes, corresponding to the three independent class labels (rumble strips, concrete barrier, and metal crash barrier). This is different from the common softmax layer whose output node values sum into one because class labels are assumed to be exclusive to each other. We use the binary cross-entropy loss. To get the final output labels for each image, we use a threshold of 0.5 on the sigmoid outputs.

In second approach, we transform the multi-label problem into multiple single-label problems. We learn three independent models corresponding to each class label. For instance, the independent model for metal crash barriers will only classify the image based on presence of metal crash barrier. Likewise, independent models for concrete barrier and rumble strips. In this design decision, the last layer is a sigmoid transformation layer with only 1 node, corresponding to either one of the three independent class labels. We use the binary cross-entropy loss. However, it would also be fine to use softmax activation function with categorical cross-entropy loss.

\section{Experimental Evaluation}\label{sec:experiment}
In this section, we compared our proposed method with baseline methods on two real world datasets in classification performance. Experiments were
conducted on a Dell workstation with Intel(R) Xeon(R) CPU E5-2687w v4@3.00GHz, 64GB main memory, and a Nvidia Quadro K6000 GPU with 2880 cores and 12GB memory. We used Keras with Tensorflow as backend to run the deep learning models. Candidate  classification methods included:

\begin{itemize}
    \item {\bf CNN only}: We used Inception-ResnetV2 CNN model on streetview images with three classes: rumble strips, concrete barriers and metal crash barriers. We added one more dense layer with 250 nodes and a $ReLU$ activation function before the final sigmoid layer with 3 nodes. 
    \item {\bf CNN-DT}: We extracted output of second last layer (with 250 nodes) from our CNN only model (Inception-ResnetV2) as feature vectors and fed it into a Decision Tree (DT) model. We used the scikit-learn package in Python. 
    \item{\bf CNN-RF}:  We extracted output of second last layer (with 250 nodes) from our CNN only model (Inception-ResnetV2) as feature vectors and fed it into a Random Forest (RF) model. We used the scikit-learn package in Python.
   % \item{\bf CNN-RF}:  We used the CNN model (Inception-ResnetV2) to extract feature vectors from second last layer (with 250 nodes) and fed it into a Random Forest (RF) model. We used the scikit-learn package in Python.

    \item {\bf CNN-sharedLSTM}: This is our proposed model to address the issue of multi-label classification using shared CNN-LSTM model for all three class labels together. 
    \item {\bf CNN-separateLSTM}: This is our proposed  model to address the issue of multi-label classification using separate single-label independent CNN-LSTM models corresponding to each class label.
\end{itemize}

 Unless specified otherwise, we used default parameters in open source tools in baseline methods.
 
\emph{Evaluation Metrics:}
To evaluate the candidate classification methods, we used precision, recall and F-score. We computed the precision, recall and F-score for all the class labels. Finally, we computed the weighted average F-score for all candidate classification methods. To calculate the weighted average, we used equation~\ref{weightedAverage},

\begin{equation} \label{weightedAverage}
Avg.F = F_{RS}\frac{RS_{n}}{RS_{n} + MCB_{n} + CB_{n} } +  F_{MCB}\frac{MCB_{n}}{RS_{n} + MCB_{n} + CB_{n} } +  F_{CB}\frac{CB_{n}}{RS_{n} + MCB_{n} + CB_{n} }
\end{equation}
where $F_{RS}$, $F_{MCB}$, and $F_{CB}$ refers to F-score of rumble strips, metal crash barriers and concrete barriers class labels respectively. Similarly, $RS_{n}$, $MCB_{n}$, and $CB_{n}$ refers to the number of image containing class labels: rumble strips, metal crash barriers and concrete barriers respectively.

% computed the product of F-score of a given class label and the ratio of number of images containing the current class label over sum of number of images for each class label. Next, we summed the above products computed for each class label to get the weighted average. 

%add equation

\subsection{Dataset Description}
To evaluate the performance of our proposed models, first we randomly selected 3,745 labeled isolated streetview images across the state of Alabama for pre-training. We used this dataset to pre-train the CNN in all baseline and proposed methods. Next, we selected different road segments within I-20 highway in Alabama to extract spatially continuous streetview images for training, validation, and test datasets. The images in extracted datasets are different from the images in pre-training dataset.  For training and validation, we selected road segment from 33\degree37'08.4"N 85\degree42'28.2"W to 33\degree35'06.5"N 86\degree03'40.6"W in I-20 East near Oxford, Alabama. We selected two test datasets. Our first test dataset was based on the road segment from 33\degree35'09.6"N 85\degree52'32.1"W to 33\degree36'56.4"N 85\degree41'25.2"W in I-20 West near Oxford, Alabama, which is closer to the training and validation datasets. Our second dataset was based on the road segment from 33\degree08'09.1"N 87\degree38'05.6"W to 33\degree11'13.2"N 87\degree19'59.3"W in I-20 West near Tuscaloosa, Alabama, which is far from the training and validation datasets. We refer to the first test dataset as Test Set\_$Oxford$ and second test dataset as Test Set\_$Tuscaloosa$ in rest of the paper. We used same training and validation dataset for both of the test datasets. 

We divided the road segments into an equally distanced set of geolocation coordinates. We set the distance interval of 20 meters. We then used Google Street View API to download the streetview images  respective to each coordinate. We have three safety features classes: rumble strips (\textbf{RS}), metal crash barriers (\textbf{MCB}), and concrete barriers (\textbf{CB}). Table ~\ref{tab:classdistribution} shows the number of images and class distribution for training, validation and test datasets.

% To evaluate the performance of the proposed model, we  selected different road segments within I-20 highway to extract continuous streetview imagery for training, validation, and testing. We selected a road segment in i-20 East for training and validation and two road segments in i-20 West for testing. The extent of training road segment is from X to Y in i-20 East whereas the extent of validation road segment is from A to B in i-20 East segment. We use same training and validation dataset for both of the test datasets. Our first test dataset is based on the road segment from X to Y in i-20 West which is closer to the training and validation dataset. Our second dataset is based on the road segment from X to Y in i-20 West. The difference between our two test datasets is that one is located closer to the training and validation data location and another lies far from the training and validation data location. We divided the road segments into an equally distanced set of geolocation coordinates. We set the distance interval of 20 meters. We then used Google Street View API to download the streetview images  respective to each coordinate. We have three safety features classes: rumble strips (\textbf{RS}), concrete barriers (\textbf{CB}) and metal crash barriers (\textbf{MCB}). Table ~\ref{tab:classdistribution} shows the number of images and class distribution for training, validation and test datasets.  

\begin{table}[h]
\centering
\caption{Class Distribution}
\label{tab:classdistribution}
\begin{tabular}{|l|l|l|l|l|}
\hline
               & \begin{tabular}[c]{@{}l@{}}Number \\ of Images\end{tabular} &\begin{tabular}[c]{@{}l@{}}Rumble \\ Strips\end{tabular} & \begin{tabular}[c]{@{}l@{}}Metal \\Crash \\ Barriers\end{tabular} & \begin{tabular}[c]{@{}l@{}}Concrete \\ Barriers\end{tabular} \\ \hline
Pre-Training Set   & 3745       & 1882            &  1149            & 1632          \\ \hline               
Training Set   & 983       & 868            & 324               & 352         \\ \hline
Validation Set & 594       & 493            & 96               &  224         \\ \hline
Test Set\_$Oxford$       & 950       & 857            &279               & 354        \\ \hline
Test Set\_$Tuscaloosa$       & 1350       & 879            & 403  & 784         \\ \hline
\end{tabular}
\end{table}

% We pre-train the CNN model with the above mentioned pre-training images. We then retrain the pre-trained CNN model with the training and validation set for all the baseline and proposed methods.

% We  pre-train CNN model with 2,750 streetview images randomly selected from road network in Alabama except the images in above mentioned training, validation and test datasets. We used this pre-train CNN model retrain with the training and validation set in all the baseline and proposed methods.
% To train our proposed CNN-LSTM models, we used the input vector length of 50 spatially continuous images for each sequence. We used a sliding window of 1 on training and validation data set to create training and validation sequences for CNN-LSTM models. We generated 883 training and 544 validation sequence. 

\subsection{Hyperparameter Settings}
For our proposed models, there are several design decision to be made including input vector length in LSTM, dimension of the hidden state of LSTM, learning rate, dropout, optimization function and training batch size. 
To train our proposed CNN-LSTM models, we used the input vector length of 50 spatially continuous images on each sequence. We used a sliding window of 1 on training and validation datasets to create training and validation sequences for our proposed CNN-LSTM models. We generated 883 training and 544 validation sequences. We set the input vector length in LSTM as 50, and the dimension of the hidden state of LSTM as 100. The input vector length as 50 refers to streetview images covering 1000 meters (20 meters separation between images). For learning rate, we first started with high learning rate of $10^{-3}$ and observed  oscillating training and validation loss curves. We then gradually decreased the learning rate and obtained more stable curves with the learning rate of $10^{-6}$ for CNN and CNN-LSTMs models. Next, we varied the dropout value from $0.2$ to $0.6$. We observed that with the higher value of dropout the convergence of training and validation loss was slower and required higher number of epochs. So, we tuned the dropout to the optimal value of $0.2$.  We trained CNN and CNN-LSTMs models using Adam optimizer. The training batch size for CNN model was 32 and that for CNN-LSTMs were 1.  
\begin{figure}[h!]
	\centering
	\subfigure{\includegraphics[width=2.5in]{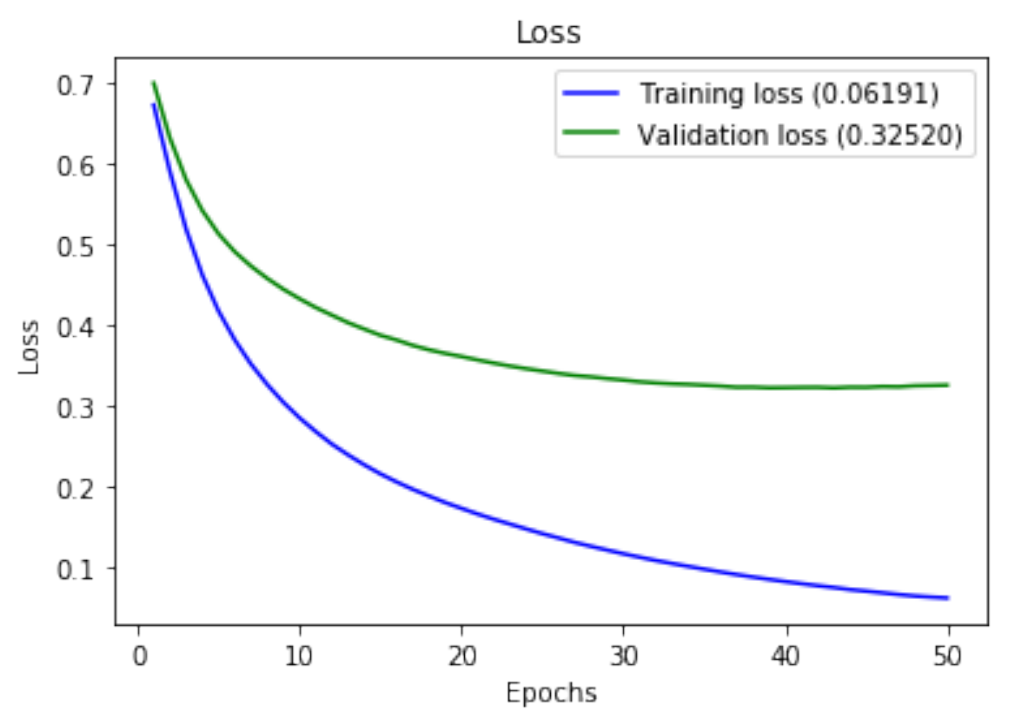}}
	\subfigure{\includegraphics[width=2.5in]{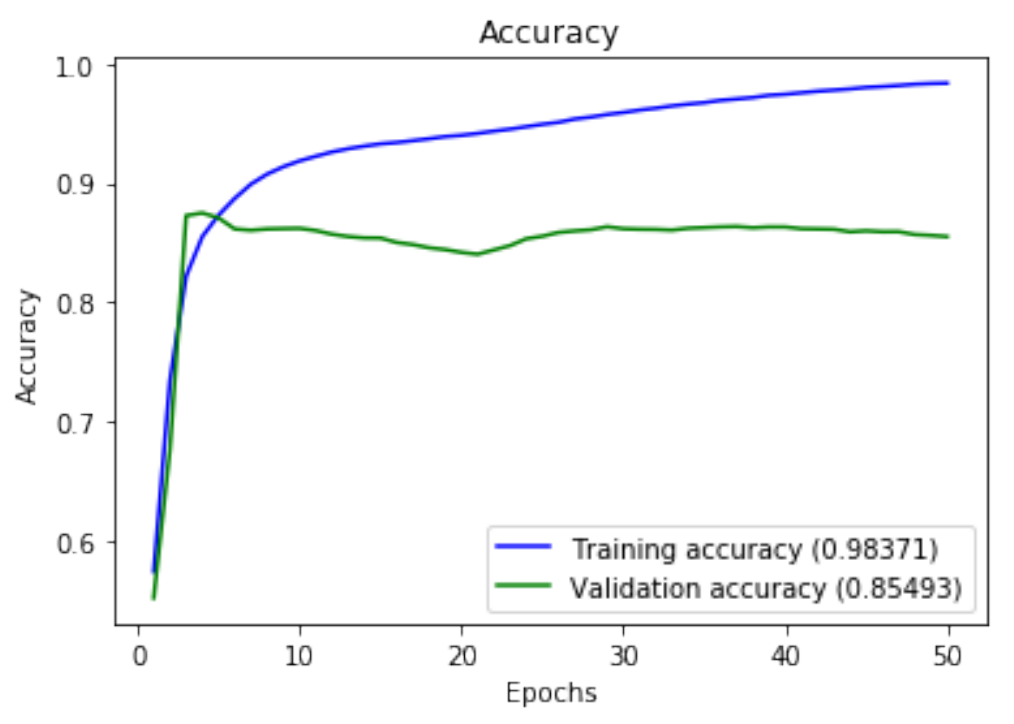}}
	\caption{Training and Validation performance of CNN-sharedLSTM over 50 epochs for all 3 class labels}
	\label{fig:learningCNNLSTM-1}
\end{figure}

\begin{figure}[h!]
	\centering
	\subfigure{\includegraphics[width=2.5in]{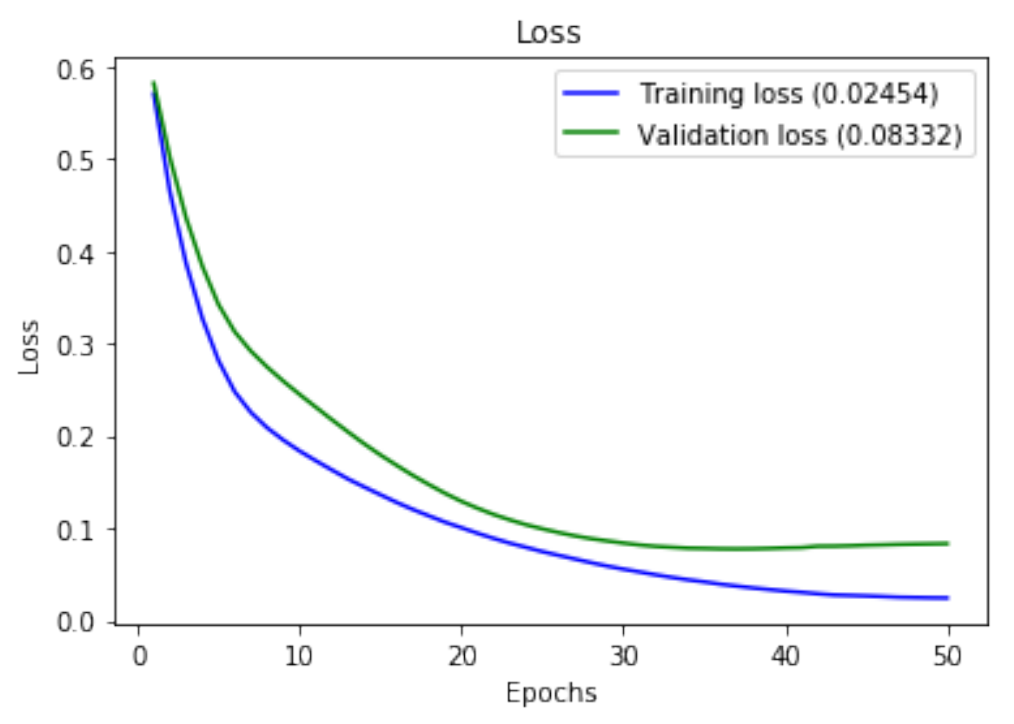}}
	\subfigure{\includegraphics[width=2.5in]{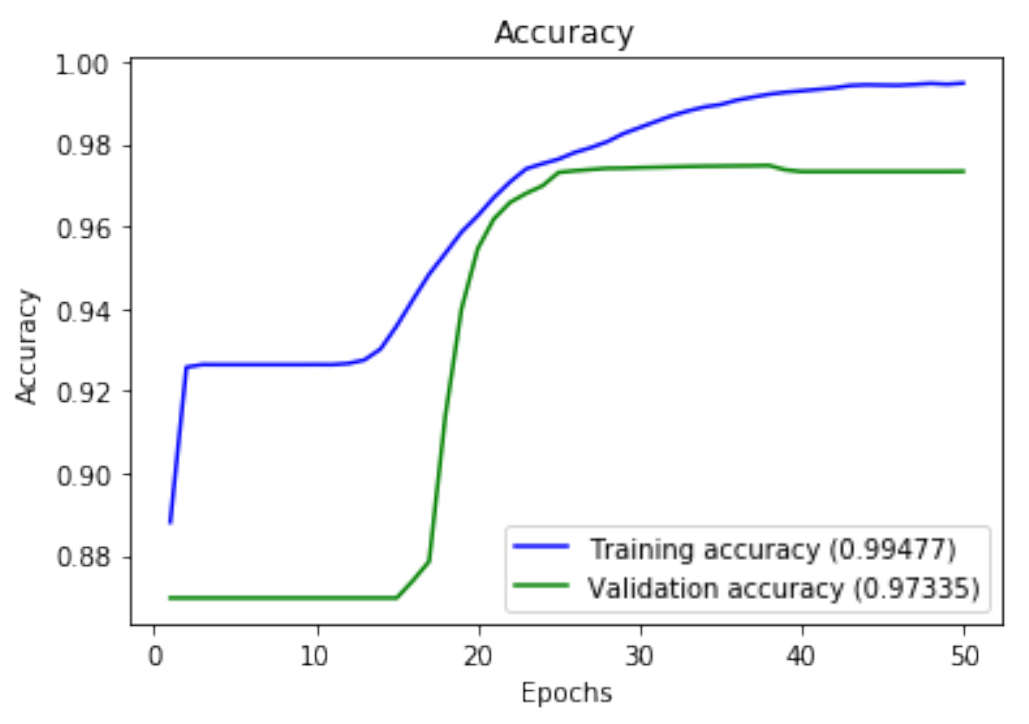}}
	\caption{Training and Validation performance of CNN-separateLSTM for rumble strips over 50 epochs}
	\label{fig:learningRS}
\end{figure}

\begin{figure}[h!]
	\centering
	\subfigure{\includegraphics[width=2.5in]{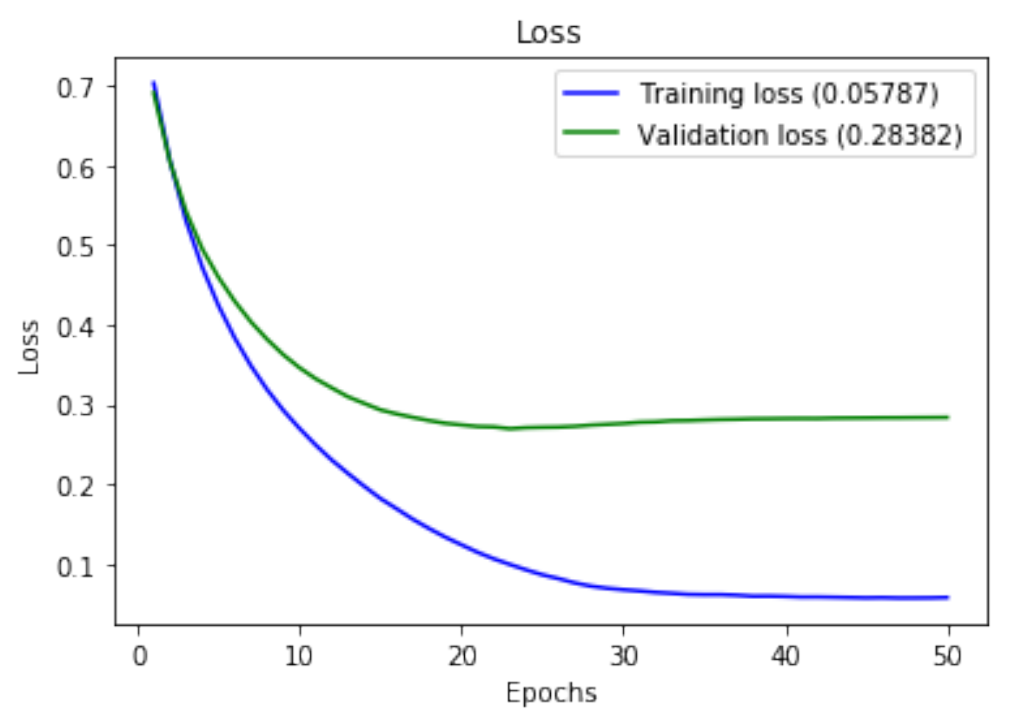}}
	\subfigure{\includegraphics[width=2.5in]{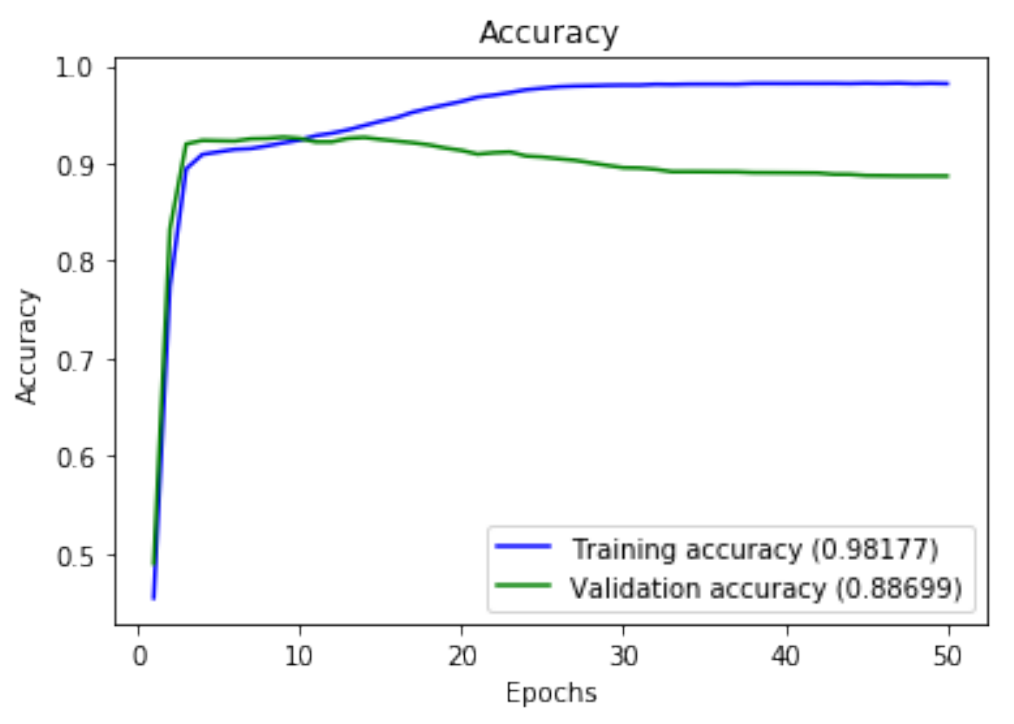}}
	\caption{Training and Validation performance of CNN-separateLSTM for metal crash barriers over 50 epochs}
	\label{fig:learningMC}
\end{figure}

\begin{figure}[h!]
	\centering
	\subfigure{\includegraphics[width=2.5in]{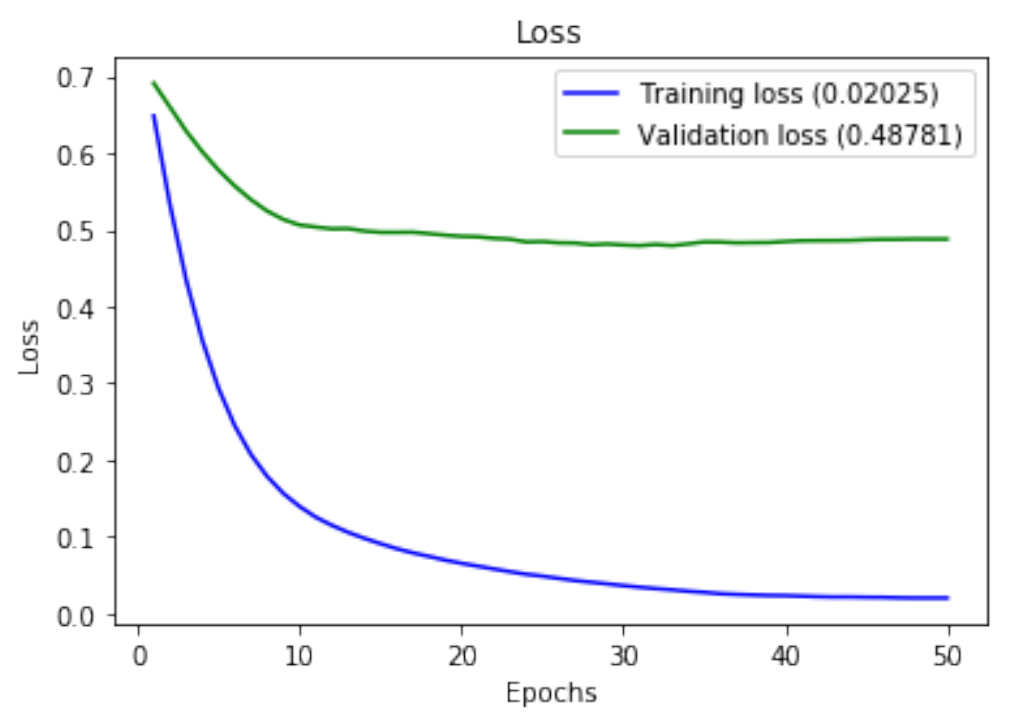}}
	\subfigure{\includegraphics[width=2.5in]{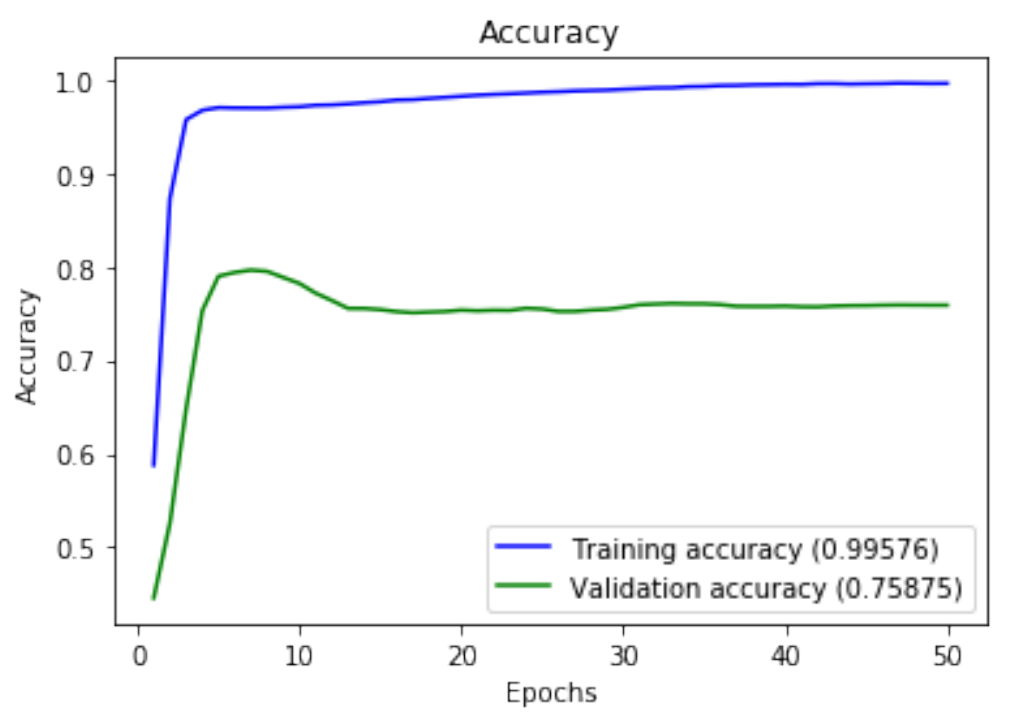}}
	\caption{Training and Validation performance of CNN-separateLSTM for concrete barriers over 50 epochs}
	\label{fig:learningCB}
\end{figure}

\subsection{Classification Performance}
Figure~\ref{fig:learningCNNLSTM-1} shows the training performance of CNN-sharedLSTM model. Similarly, Figure~\ref{fig:learningRS}, ~\ref{fig:learningMC}, and ~\ref{fig:learningCB} shows the training performance of CNN-separateLSTM model for rumble strips, metal crash barriers, and concrete barriers class labels respectively. The training and validation loss for CNN-sharedLSTM are 0.06 and 0.32 respectively.  In case of CNN-separateLSTM, the training and validation loss for rumble strips class label are 0.02 and 0.08 respectively. We can observe that the gap between training and validation loss for rumble strips in CNN-separateLSTM model is lower than that of CNN-sharedLSTM model. The training and validation loss in CNN-separateLSTM for metal crash barriers class label are 0.05 and 0.28 respectively, which also shows similar trend. It is because the CNN-separateLSTM model can better learn the spatial scale of rumble strips and metal crash barrier separately. Rumble stirps have very high spatial scale in comparison to the metal crash barriers. Furthermore, rumble strips and metal crash barriers are mostly easy to identify in the streetview images. Finally, the training and validation loss for concrete barrier class label are 0.02 and 0.48 respectively. As shown in in Figure~\ref{fig:learningCB}, the gap between training and validation curve is high and indicates over-fitting. It is likely because the concrete barriers are very diverse and hard to learn. We observe the texture of concrete barriers in some regions are very similar to the texture of the roads. 

% We achieved training accuracy of around 0.98 and validation accuracy of 0.85.
\begin{table}
\centering
\caption{Classification on Test Set\_$Oxford$}
\label{tab:results1}
\begin{tabular}{cccccc}
\hline
Classifiers & Class & Precision &Recall & F & Avg. F\\ \hline
%section1: non-spatial ensemble
\multirow{3}{*}{CNN-DT}&RS&{0.95}&{0.83}&{0.89}&\multirow{3}{*}{0.85}\\ 
 &MCB&{0.79}&{0.82}&{0.80}&\\
 &CB&{0.81}&{0.76}&{0.78}&\\ 
  \hline
 
 \multirow{3}{*}{CNN-RF}&RS&{0.94}&{0.87}&{0.90}&\multirow{3}{*}{0.87}\\
  &MCB&{0.89}&{0.86}&{0.87}&\\ 
 &CB&{0.78}&{0.82}&{0.80}&\\ 
 \hline
 
 \multirow{3}{*}{CNN only}&RS&{0.94}&{0.95}&{0.95}&\multirow{3}{*}{0.89}\\
  &MCB&{0.91}&{0.78}&{0.84}&\\
 &CB&{0.74}&{0.85}&{0.79}&\\ 
  \hline
 
  \multirow{3}{*}{CNN-sharedLSTM}&RS&{0.43}&{0.98}&{0.96}&\multirow{3}{*}{0.91}\\ 
   &MCB&{0.90}&{0.83}&{0.86}&\\
 &CB&{0.74}&{0.91}&{0.82}&\\ 
  \hline
 
  \multirow{3}{*}{CNN-separateLSTM}&RS&{0.95}&{0.96}&{0.96}&\multirow{3}{*}{0.92}\\ 
   &MCB&{0.93}&{0.83}&{0.88}&\\
 &CB&{0.77}&{0.93}&{0.84}&\\ 
  \hline

\end{tabular}
\end{table}

\begin{table}
\centering
\caption{Classification on Test Set\_$Tuscaloosa$}
\label{tab:results2}
\begin{tabular}{cccccc}
\hline
Classifiers & Class & Precision &Recall & F & Avg. F\\ \hline
%section1: non-spatial ensemble
\multirow{3}{*}{CNN-DT}&RS&{0.85}&{0.84}&{0.85}&\multirow{3}{*}{0.75}\\ 
 &MCB&{0.50}&{0.86}&{0.63}&\\
 &CB&{0.94}&{0.54}&{0.69}&\\ 
  \hline
 
 \multirow{3}{*}{CNN-RF}&RS&{0.88}&{0.87}&{0.88}&\multirow{3}{*}{0.77}\\ 
  &MCB&{0.58}&{0.77}&{0.66}&\\
 &CB&{0.93}&{0.58}&{0.71}&\\ 
  \hline
 
 \multirow{3}{*}{CNN only}&RS&{0.87}&{0.97}&{0.92}&\multirow{3}{*}{0.78}\\ 
 &MCB&{0.78}&{0.73}&{0.75}&\\ 
 &CB&{0.89}&{0.49}&{0.63}&\\ 
  \hline
 
  \multirow{3}{*}{CNN-sharedLSTM}&RS&{0.85}&{0.98}&{0.91}&\multirow{3}{*}{0.79}\\
   &MCB&{0.68}&{0.84}&{0.75}&\\ 
 &CB&{0.93}&{0.53}&{0.68}&\\ 
 \hline
 
   \multirow{3}{*}{CNN-separateLSTM}&RS&{0.88}&{0.96}&{0.92}&\multirow{3}{*}{0.83}\\ 
    &MCB&{0.74}&{0.80}&{0.77}&\\
 &CB&{0.96}&{0.61}&{0.76}&\\ 
  \hline

\end{tabular}
\end{table}
We compare different candidate methods on precision, recall, and F-score. To obtain the predicted class label, we set the probability threshold of 0.5. A road safety feature probability value above the threshold indicates presence of the road safety feature in the image. Results are summarized in Table~\ref{tab:results1} and ~\ref{tab:results2} for Test Set\_$Oxford$ and Test Set\_$Tuscaloosa$ respectively. 

For Test Set\_$Oxford$ in Table~\ref{tab:results1}, the average F-score of CNN-DT, CNN-RF and CNN only models are 0.85, 0.87 and 0.89 respectively. we can observe the average F-score for CNN with DT and  RF is lower than CNN only. It is because CNN-DT and CNN-RF takes the output of 2nd last layer (with 250 output dimension) of CNN only model as the features and fits the models. But the last layer in CNN only model, a dense layer with 3 nodes, have extra learnable parameters which can help CNN only model train better. Next, we also observe that both of our proposed models: CNN-sharedLSTM and CNN-separateLSTM perform better than CNN only model with average F-score of 0.91 and 0.92 respectively. CNN only model may fail to correctly identify some inbetween images in the test image sequence. Our proposed models can correct these errors by incorporating spatial dependency in the learning process using LSTM network. Also, we can observe that our CNN-separateLSTM performs better than CNN-sharedLSTM. It is because in case of CNN-separateLSTM, the independent models for each class label can better learn the spatial scale of different class labels separately. For example, metal crash barrier have an average length of few hundred meters wheres other road safety features such as rumble strips may have average length of few kilometers. Furthermore, for CNN-separateLSTM, we can observe very high F-score of 0.96 on rumble strips class  which is consistent with the learning curve in Figure~\ref{fig:learningRS} with small gap between the training and validation loss. In case of metal crash barriers, we observe F-score of 0.88 which is also consistent with the learning curve shown in Figure~\ref{fig:learningMC} with medium gap between training and validation loss. Finally, for concrete barrier, we observe lowest F-score of 0.84, also consistent with the learning curve shown in Figure~\ref{fig:learningCB} with high gap between training and validation loss. 
  \begin{figure*}[h]
  \centering
	\includegraphics[width=5.6in]{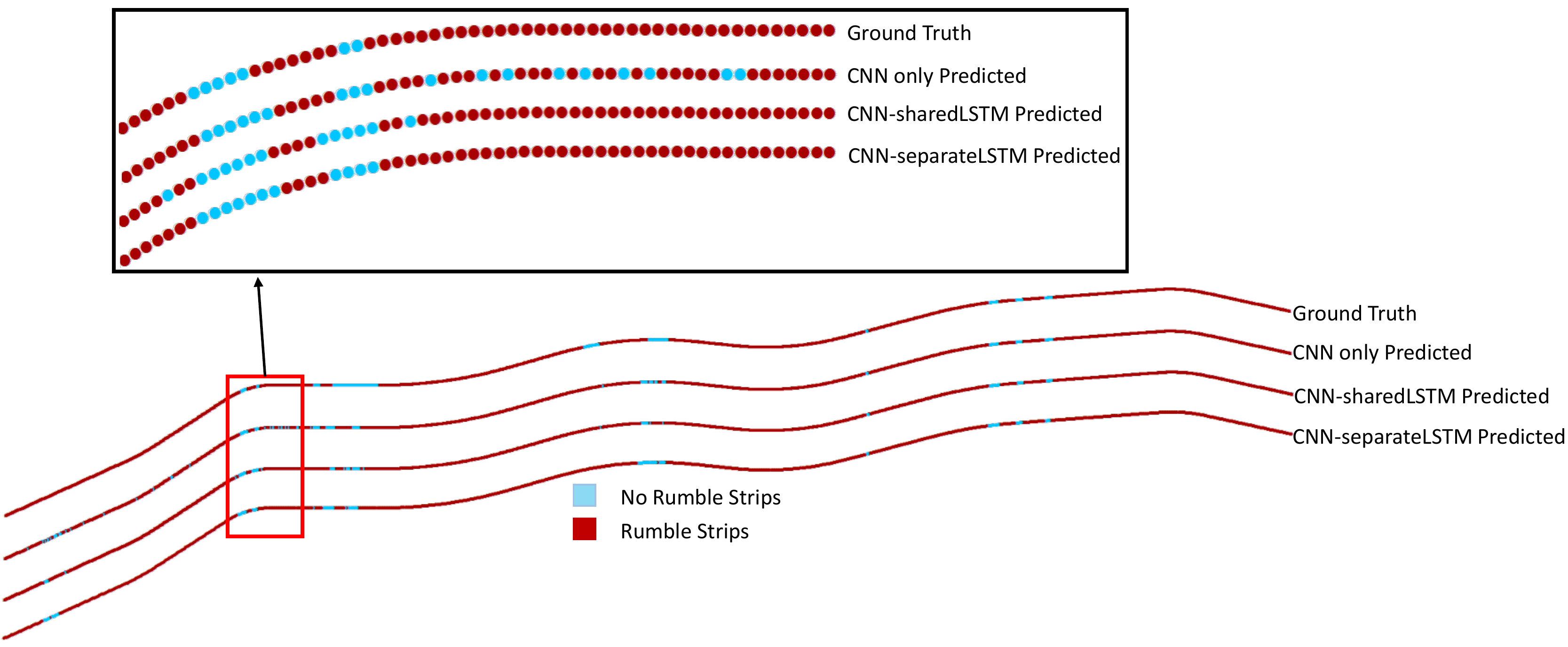}
	\caption{Prediction maps on Test Set\_$Oxford$ for rumble strips}
	\label{fig:rsmapping}
\end{figure*}

  \begin{figure*}[h]
    \centering
	\includegraphics[width=5.5in]{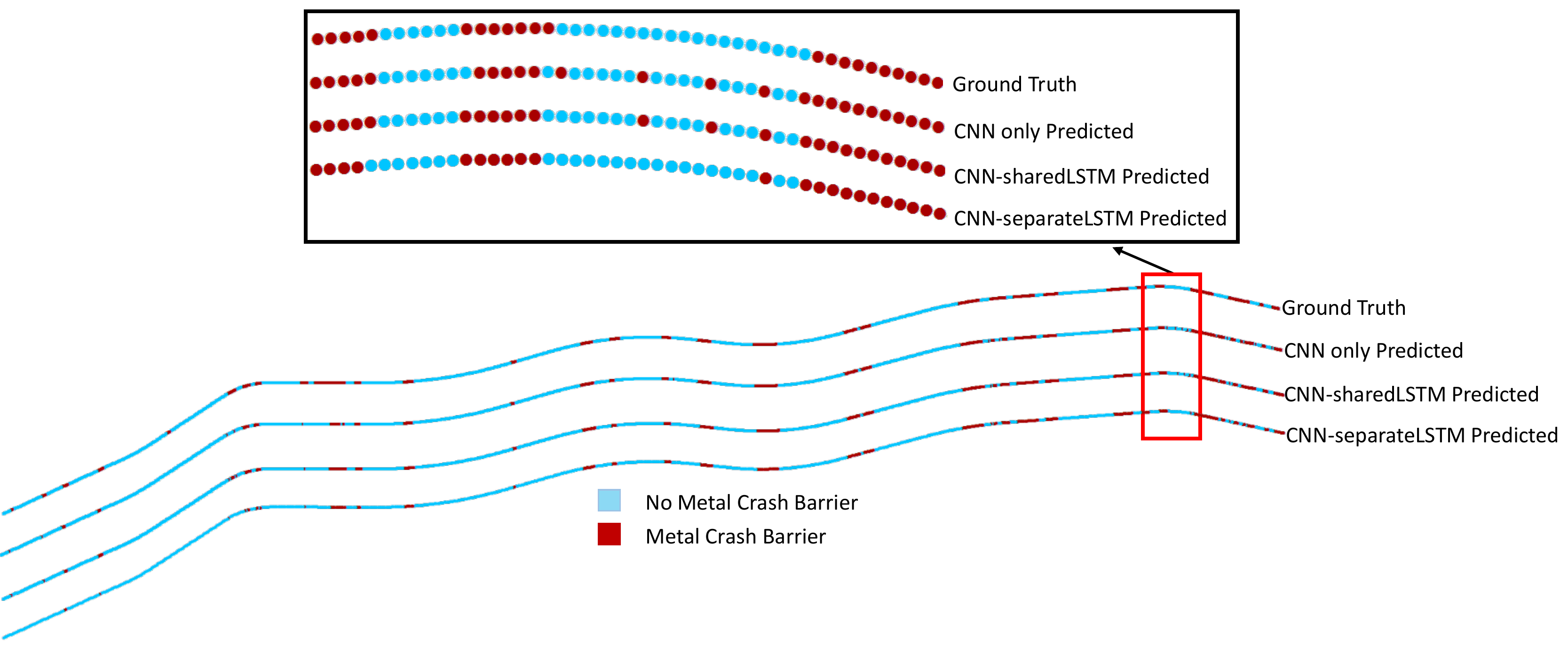}
	\caption{Prediction maps on Test Set\_$Oxford$ for metal crash barriers}
	\label{fig:mcbmapping}
\end{figure*}

  \begin{figure*}[h]
    \centering
	\includegraphics[width=5.5in]{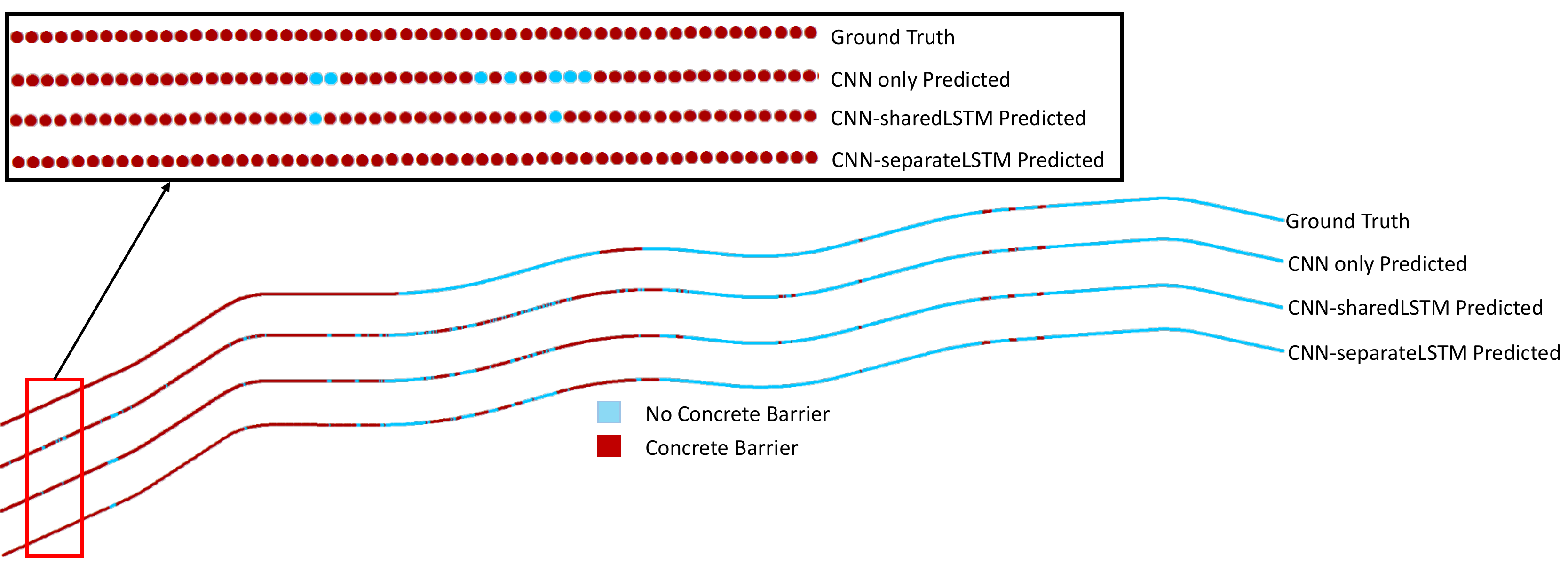}
	\caption{Prediction maps on Test Set\_$Oxford$ for concrete barriers}
	\label{fig:cbmapping}
\end{figure*}

\begin{figure*}[h]
  \centering
	\includegraphics[width=5.6in]{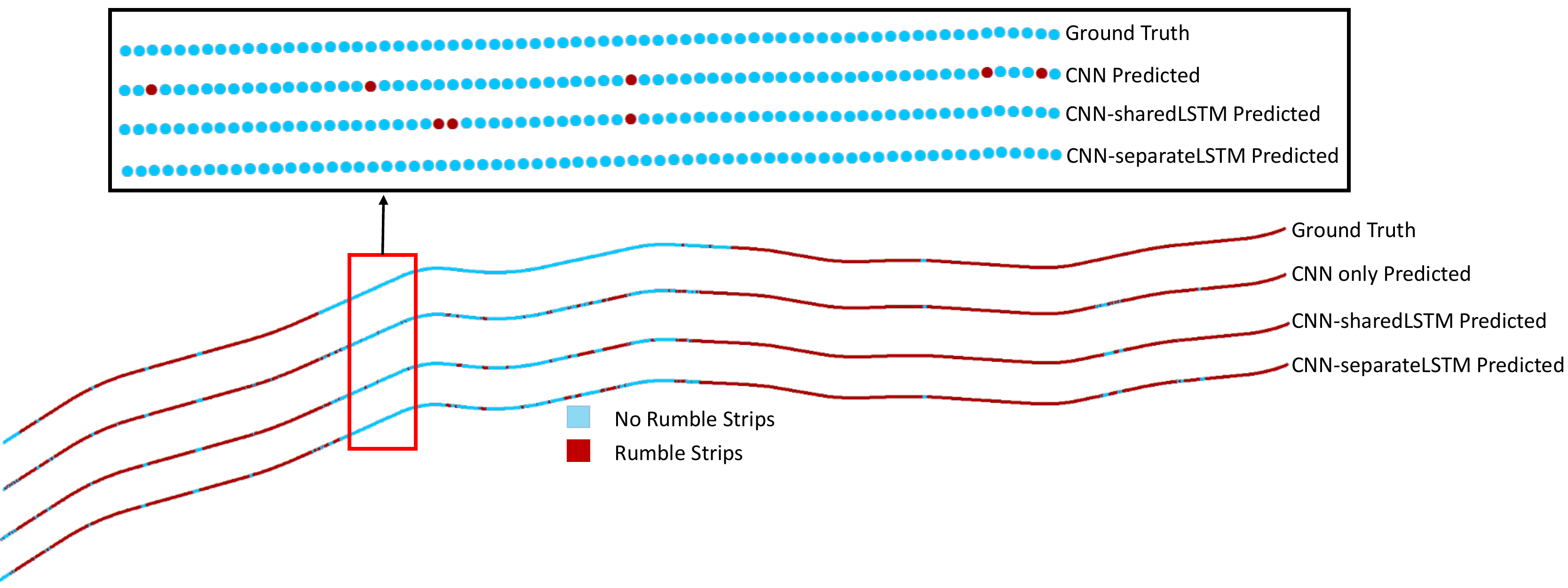}
	\caption{Prediction maps on Test Set\_$Tuscaloosa$ for rumble strips}
	\label{fig:rsmapping2}
\end{figure*}

  \begin{figure*}[h]
  \centering
	\includegraphics[width=5.5in]{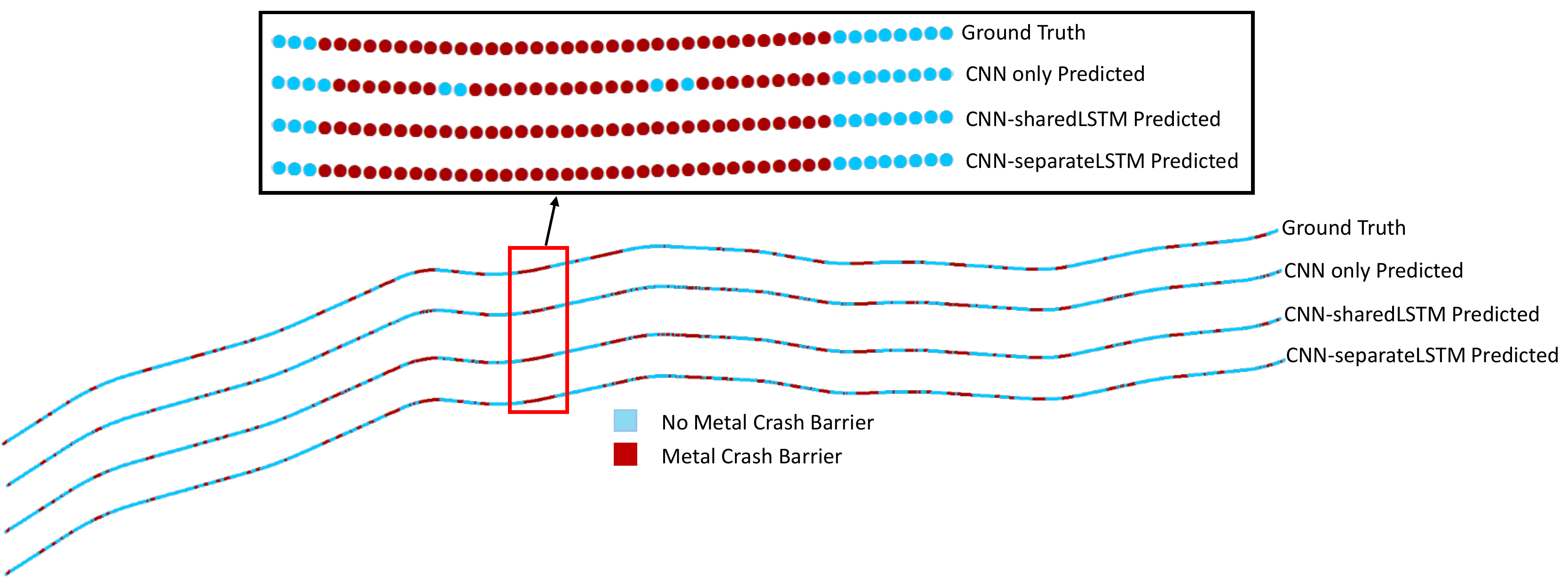}
	\caption{Prediction maps on Test Set\_$Tuscaloosa$ for metal crash barriers}
	\label{fig:mcbmapping2}
\end{figure*}

  \begin{figure*}[h]
  \centering
	\includegraphics[width=5.6in]{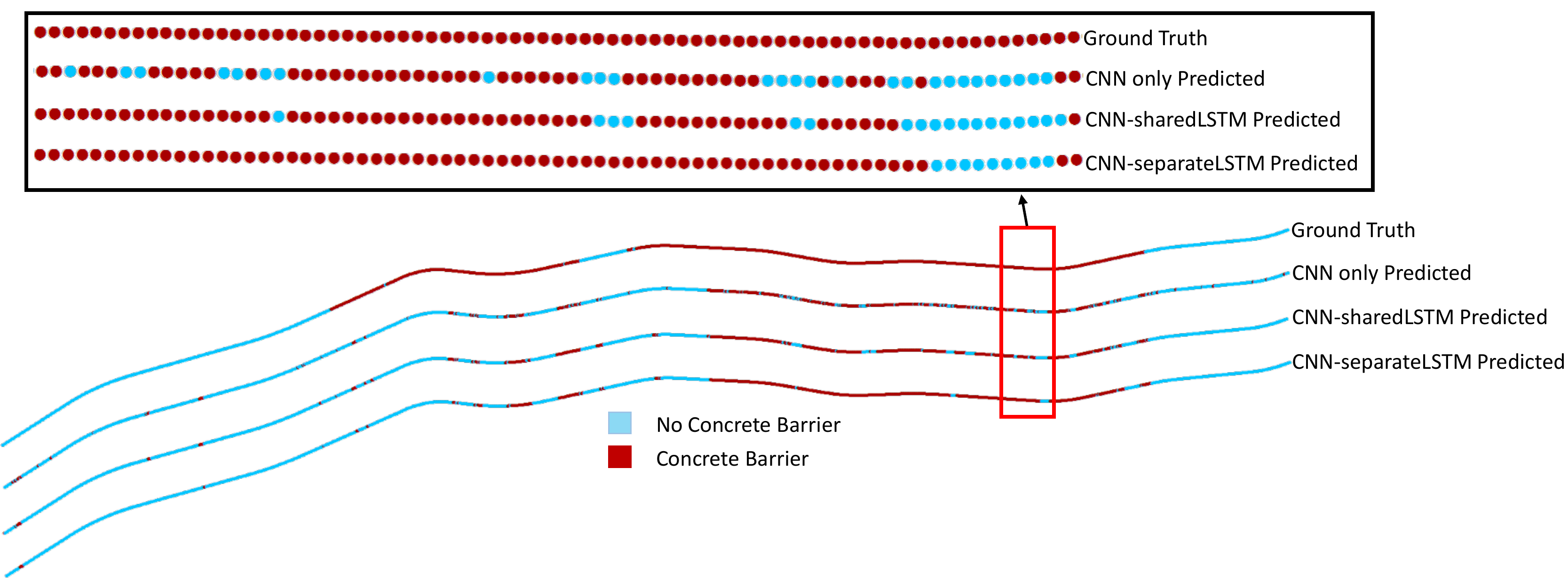}
	\caption{Prediction maps on Test Set\_$Tuscaloosa$ for concrete barriers}
	\label{fig:cbmapping2}
\end{figure*}

For Test Set\_$Tuscaloosa$ in Table~\ref{tab:results2}, we can also observe similar trends as discussed above. But, the overall performance of all the candidate methods is lower. The average F-score of CNN-DT, CNN-RF, CNN only, CNN-sharedLSTM, and CNN-separateLSTM are 0.75, 0.77, 0.78, 0.79, and 0.83 respectively. It is likely because the Test Set\_$Tuscaloosa$ image location lies far from the training and validation region. So, the training and validation images might not be very similar to images in Test Set\_$Tuscaloosa$. It can be easily improve by introducing some representative images closer to Test Set\_$Tuscaloosa$ in training and validation data.

We also visualize the predicted class label on map based on all the predicted results summarized in Table~\ref{tab:results1} and Table~\ref{tab:results2}.
Figures~\ref{fig:rsmapping}, ~\ref{fig:mcbmapping} and ~\ref{fig:cbmapping} shows the ground truth and prediction maps for three road safety feature classes based on CNN only, CNN-sharedLSTM and CNN-separateLSTM models on Test Set\_$Oxford$. Similarly, Figures~\ref{fig:rsmapping2}, ~\ref{fig:cbmapping2} and ~\ref{fig:mcbmapping2} shows ground truth and prediction maps on Test Set\_$Tuscaloosa$.
%groundtruth
We can observe that different road safety features have different spatial scale. From Figure~\ref{fig:rsmapping} and ~\ref{fig:rsmapping2}, we can observe that ground truth spatial scale of rumble strips are usually very long. From Figure~\ref{fig:mcbmapping} and ~\ref{fig:mcbmapping2}, we can observe that ground truth spatial scale of metal crash barriers are usually very short. And finally, from Figure~\ref{fig:cbmapping} and ~\ref{fig:cbmapping2}, we can observe that ground truth spatial scale of concrete barriers can vary from short to long. Next, in all the prediction maps, we can observe CNN only based predictions contain some isolated errors. But, CNN-sharedLSTM model was able to correct some of those isolated errors as highlighted in zoom-in sub-visualizations. However, we observer CNN-separateLSTM model to be more accurate, which is consistent with the summarized result in Table~\ref{tab:results1} and ~\ref{tab:results2}. The CNN-sharedLSTM and CNN-separateLSTM models are able to generate more accurate map than CNN only model due to the incorporation of spatial sequential structure in the learning process.  

\begin{figure*}[h]
\centering
	\includegraphics[width=5.5in]{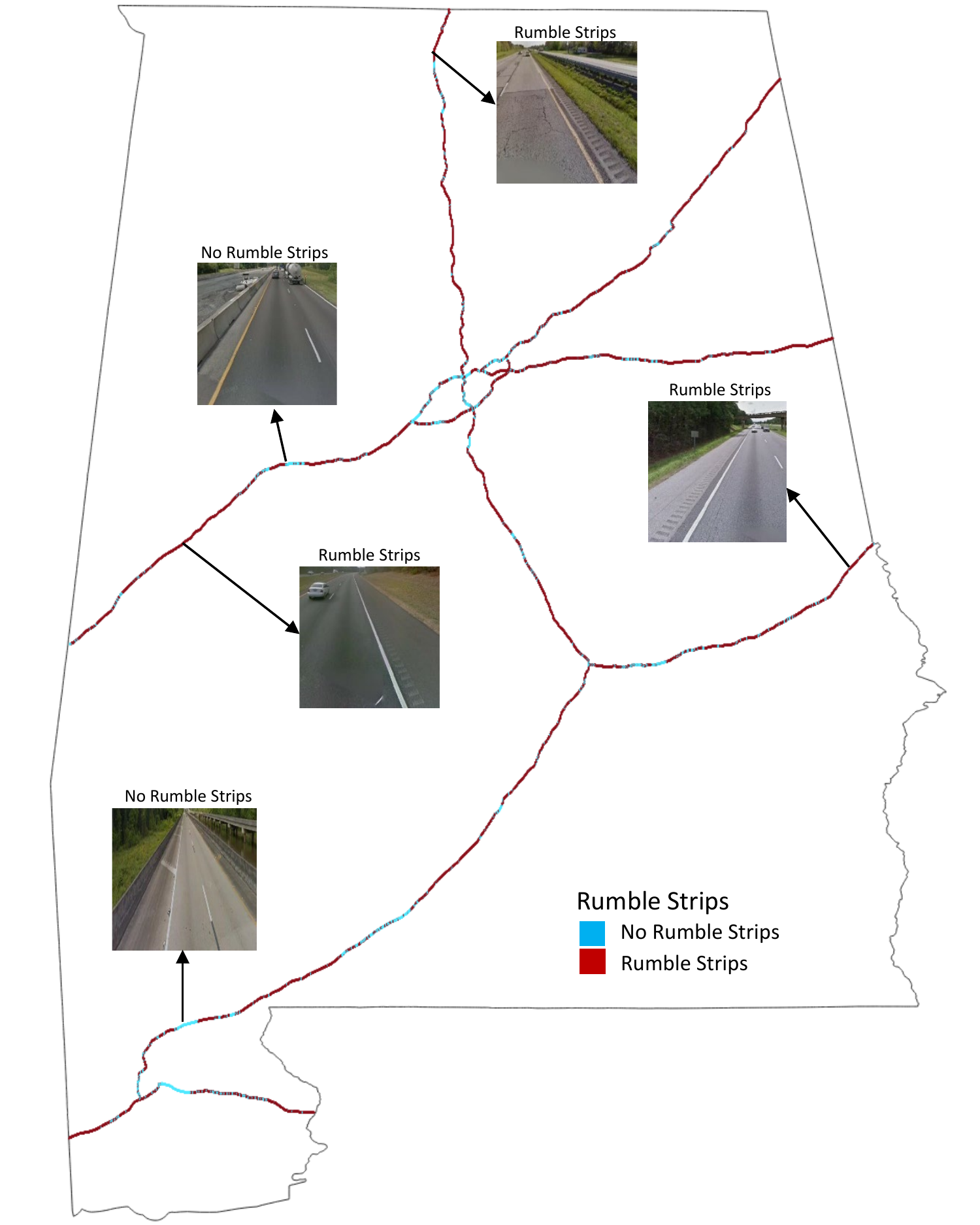}
	\caption{Classification map for rumble strips class label using CNN-separateLSTM model}
	\label{fig:rsmap}
\end{figure*}

 \begin{figure*}[h]
 \centering
	\includegraphics[width=5.5in]{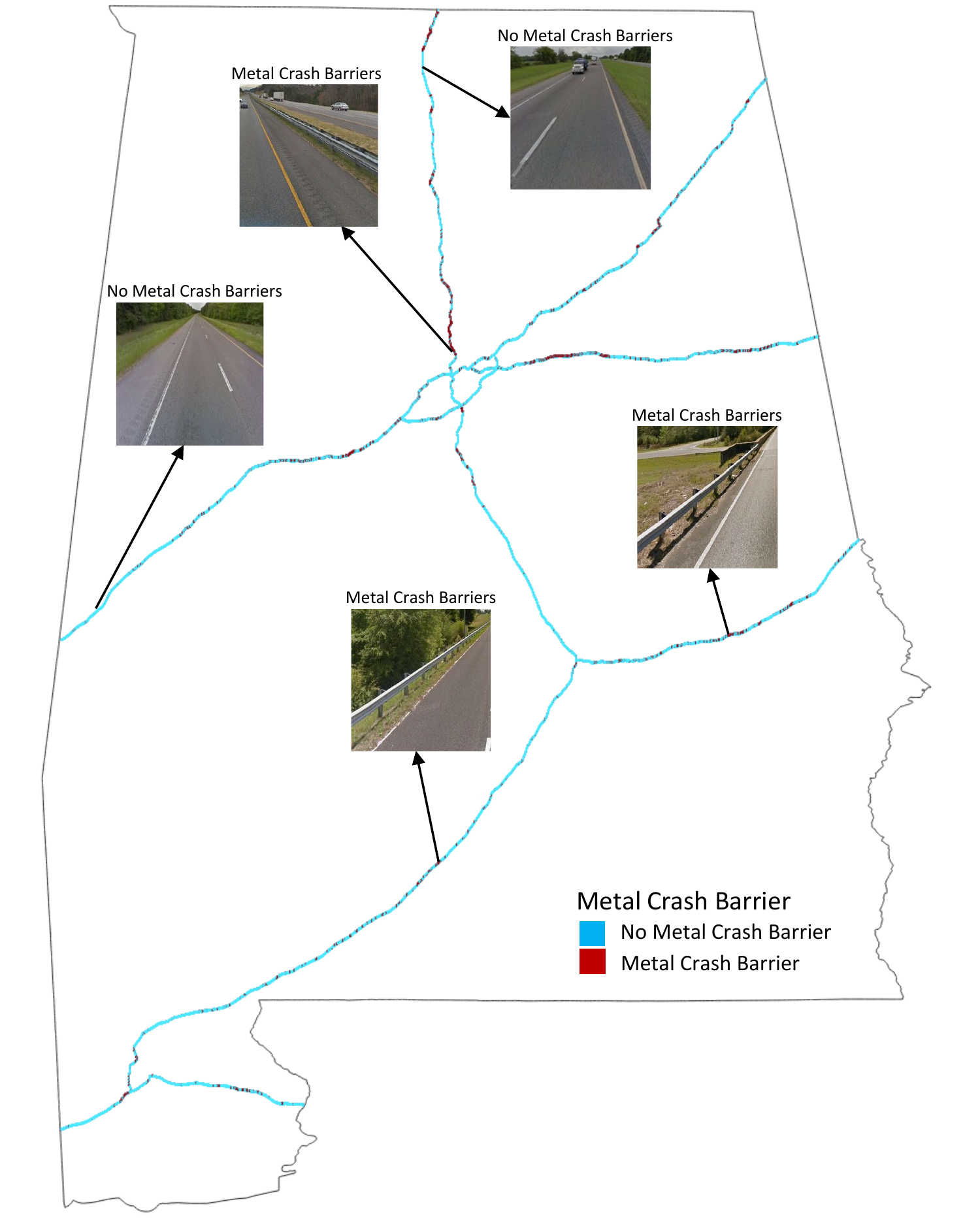}
	\caption{Classification map for metal crash barriers class label using CNN-separateLSTM model}
	\label{fig:mcmap}
\end{figure*}

 \begin{figure*}[h]
 \centering
	\includegraphics[width=5.5in]{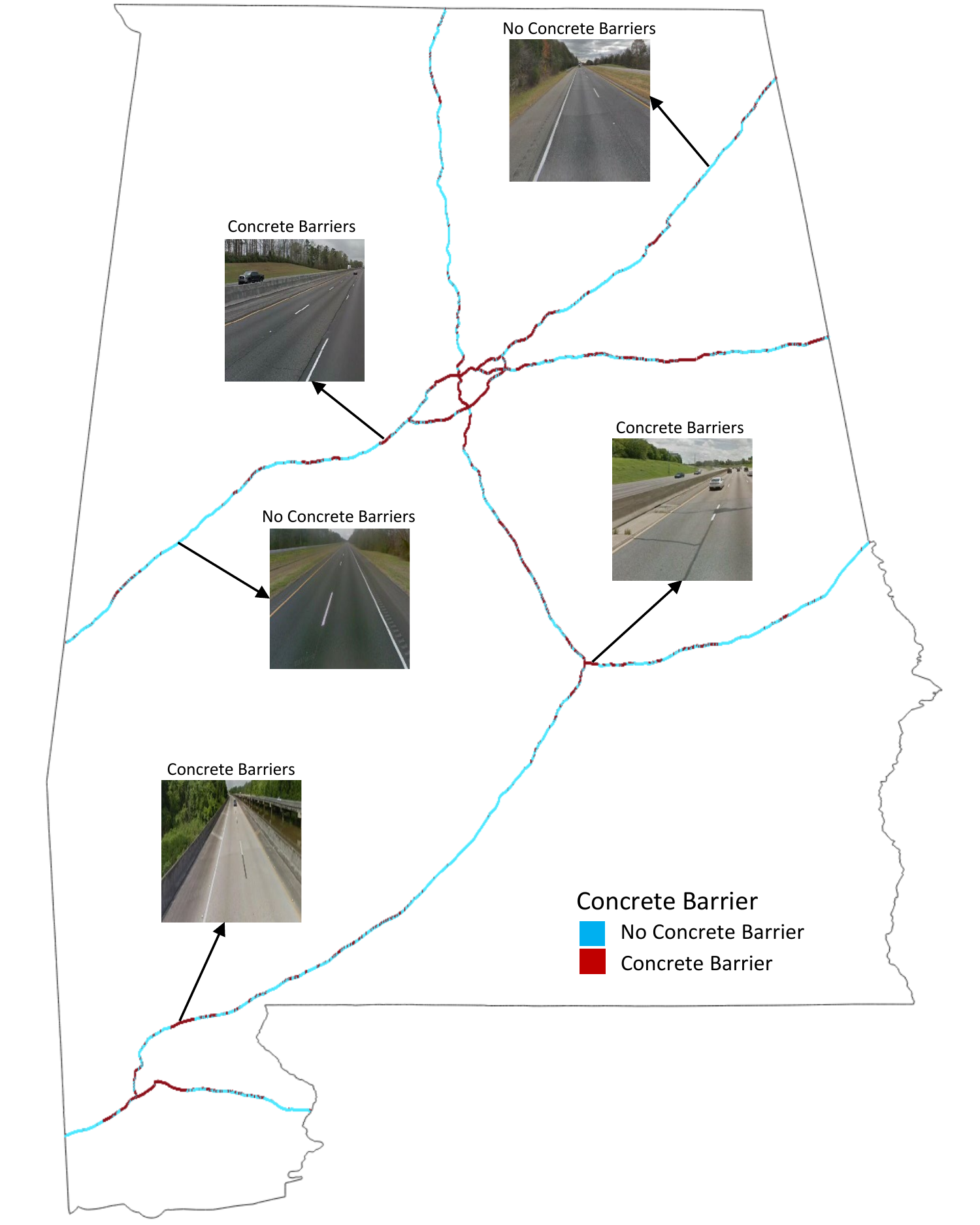}
	\caption{Classification map for concrete barriers class label using CNN-separateLSTM model}
	\label{fig:cbmap}
\end{figure*}
% past tense 
\subsection{Case Study}
We downloaded 69,500 streetview images of all the major highways in the entire state of Alabama. The major highways include I-10, I-20, I-59, I-65, I-85 and I-459 within Alabama. We then classified the road safety features in those images using our CNN-separateLSTM model. The map for each road safety features based on the classification results is shown in figures ~\ref{fig:rsmap}, ~\ref{fig:mcmap} and ~\ref{fig:cbmap}. We randomly checked few data points to verify the classification result and found that most of the images were accurately classified. 

%Add summary statistics
According to the predicted maps, 54,890 streetview images were classified to have rumble strips which is around 1200 km out of 1390 km in major highways in Alabama. 12,841 streetview images were classified to have metal crash barriers which implies around 257 km out of 1390 km. Finally, 26,997 streetview images were classified to have concrete barriers which implies around 540 km out of 1390 km.
We observed that the spatial scale of the predicted map for rumble strips are very long which is consistent with the spatial scale observed in the ground truth for test datasets. Similarly, from the predicted map in Figure~\ref{fig:mcmap}, we observed that the metal crash barriers are evenly distributed throughout the major highways in Alabama and have short spatial scale. It is also consistent with the test datasets. Finally, from the predicted  map in Figure~\ref{fig:cbmap}, we observe that the spatial scale of concrete barriers can vary. We further observed that the long chain concrete barriers are usually located near city areas. Short concrete barriers are usually placed in the bridges. Also, we noticed that the bridges with the concrete barriers usually do not contain rumble strips.    

 \section{Conclusion}\label{sec:conclusion}
In this paper, we proposed two different CNN-LSTM based spatial classification models for mapping safety features along road networks.  Our CNN-lSTM models can capture spatial linear structure between consecutive images along a road network path. Results on real world Google Streetview images collected in Alabama showed that our models outperform several baseline methods. Furthermore, through experimental evaluation, we found out that the separate CNN-LSTM models for independent class labels performed better than shared CNN-LSTM model.

In future work, we plan to investigate more general spatial network structure with graph-LSTM. 

\section{Acknowledgement}
This material is based upon work supported by Alabama Transaportation Institute.

\bibliographystyle{unsrt}  
\bibliography{references}  %%% Remove comment to use the external .bib file (using bibtex).


% George Kour and Raid Saabne.
% \newblock Real-time segmentation of on-line handwritten arabic script.
% \newblock In {\em Frontiers in Handwriting Recognition (ICFHR), 2014 14th
%   International Conference on}, pages 417--422. IEEE, 2014.

% George Kour and Raid Saabne.
% \newblock Fast classification of handwritten on-line arabic characters.
% \newblock In {\em Soft Computing and Pattern Recognition (SoCPaR), 2014 6th
%   International Conference of}, pages 312--318. IEEE, 2014.

% \bibitem{hadash2018estimate}
% Guy Hadash, Einat Kermany, Boaz Carmeli, Ofer Lavi, George Kour, and Alon
% \newblock Estimate and replace: A novel approach to integrating deep neural
%   networks with existing applications.
% \newblock {\em arXiv preprint arXiv:1804.09028}, 2018.

% \end{thebibliography}

\end{document}